\documentclass[10pt,twocolumn,letterpaper]{article}

\usepackage{cvpr}
\usepackage{times}
\usepackage{epsfig}
\usepackage{graphicx}
\usepackage{amsmath}
\usepackage{amssymb}
\usepackage{capt-of}
\usepackage{subfigure}

\cvprfinalcopy % *** Uncomment this line for the final submission

\begin{document}

%%%%%%%%% TITLE
\title{Structured Control Nets for Deep Reinforcement Learning}

\author{Mario Srouji$^{*,1,2}$, Jian Zhang$^{*,1}$, Ruslan Salakhutdinov$^{1,2}$ \\
  $^*$Equal Contribution.\\
  $^1$Apple Inc., 1 Infinite Loop, Cupertino, CA 95014, USA.\\
  $^2$Carnegie Mellon University, 5000 Forbes Ave, Pittsburgh, PA 15213, USA.\\
  {\tt\small msrouji@andrew.cmu.edu \qquad \{jianz,rsalakhutdinov\}@apple.com}
 }

\maketitle

\begin{abstract}

In recent years, Deep Reinforcement Learning has made impressive advances in solving several important benchmark problems for sequential decision making. Many control applications use a generic multilayer perceptron (MLP) for non-vision parts of the policy network. In this work, we propose a new neural network architecture for the policy network representation that is simple yet effective. The proposed Structured Control Net (SCN) splits the generic MLP into two separate sub-modules: a nonlinear control module and a linear control module. Intuitively, the nonlinear control is for forward-looking and global control, while the linear control stabilizes the local dynamics around the residual of global control. We hypothesize that this will bring together the benefits of both linear and nonlinear policies: improve training sample efficiency, final episodic reward, and generalization of learned policy, while requiring a smaller network and being generally applicable to different training methods. We validated our hypothesis with competitive results on simulations from OpenAI MuJoCo, Roboschool, Atari, and a custom 2D urban driving environment, with various ablation and generalization tests, trained with multiple black-box and policy gradient training methods. The proposed architecture has the potential to improve upon broader control tasks by incorporating problem specific priors into the architecture. As a case study, we demonstrate much improved performance for locomotion tasks by emulating the biological central pattern generators (CPGs) as the nonlinear part of the architecture.

\end{abstract}

\section{Introduction}
\label{intro}

Sequential decision making is crucial for intelligent system to interact with the world successfully and optimally. In recent years, Deep Reinforcement Learning (DRL) has made significant progress on solving several important benchmark problems for sequential decision making, such as Atari \cite{mnih2015human}, Game of Go \cite{silver2017mastering}, high-dimensional continuous control simulations \cite{schulman2017proximal,lillicrap2015continuous}, and robotics \cite{levine2016end}. Many successful applications, especially control problems, still use generic multilayer perceptron (MLP) for non-vision part of the policy network. There have been only few efforts exploring adding specific structures to DRL policy network as an inductive bias for improving training sampling efficiency, episodic reward, generalization, and robustness~\cite{wang2016dueling}. 

In this work, we focus on an alternative but complementary area by introducing a policy network architecture that is simple yet effective for control problems (Figure~\ref{fig:envs}). 
Specifically, to improve the policy network architecture, we introduce control-specific priors on the structure of the policy network. 
The proposed Structured Control Net (SCN) splits the generic multilayer perceptron (MLP) into two separate sub-modules: a nonlinear control module and a linear control module. The two streams combine additively into the final action.
This approach has the benefit of being easily combined with many existing DRL algorithms.

\begin{figure}[t]
\begin{center}
\includegraphics[trim =0mm 0mm 0mm 0mm, clip,width=1.0\columnwidth]{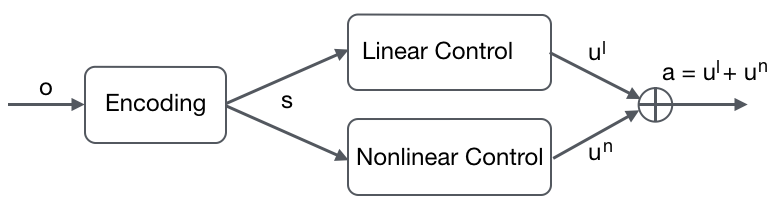}
\caption{\small The proposed Structured Control Net (SCN) for policy network that incorporates a nonlinear control module, $u^n$, and a linear control module, $u^l$. $o$ is the observation, $s$ is encoded state, and $a$ is the action output of the policy network. Here time $t$ is dropped for notation compactness. } \label{fig:arch1}
\end{center}
\vspace{-0.2in}
\end{figure}

We experimentally demonstrate that this architecture brings together the benefits of both linear and nonlinear policies by improving training sampling efficiency, final episodic reward, and generalization of learned policy, while using a smaller network compared to standard MLP baselines.
We further validate our architecture with competitive results on simulations from OpenAI MuJoCo~\cite{brockman2016openai, todorov2012mujoco}, Roboschool~\cite{roboschool}, Atari~\cite{bellemare2013arcade}, and a custom 2D urban driving environment. Our experiments are designed and conducted following the guidelines introduced by recent studies on DRL reproduciblity~\cite{henderson2017deep}. To show the general applicability of the proposed architecture, we conduct ablation tests using multiple black-box and policy gradient training methods, such as Evolutionary Strategies (ES) \cite{salimans2017evolution}, Proximal Policy Optimization (PPO)~\cite{schulman2017proximal}, and ACKTR~\cite{wu2017scalable}. 
Without any special treatment of different training methods, our ablation tests show that sub-modules of the architecture are effectively learned, and generalization capabilities are improved, compared to the standard MLP policy networks. This structured policy network with nonlinear and linear modules has the potential to be expanded to broader sequential decision making tasks, by incorporating problem specific priors into the architecture.
\section{Related Work}
\label{related}

Discrete control problems, such as Atari games, and high-dimensional continuous control problems are some of the most popular and important benchmarks for sequential decision making. They have applications in video game intelligence  \cite{mnih2015human}, simulations \cite{schulman2017proximal}, robotics \cite{levine2016end}, and self-driving cars \cite{shalev2016long}. Those problems are challenging for traditional methods, due to delayed rewards, unknown transition models, and high dimensionality of the state space. %/degrees of freedom.

Recent advances in Deep Reinforcement Learning (DRL) hold great promise for learning to solve challenging continuous control problems. Popular approaches are Evolutionary Strategy (ES) \cite{salimans2017evolution, such2017deep,conti2017improving} and various policy gradient methods, such as TRPO \cite{schulman2015trust}, A3C \cite{mnih2016asynchronous}, DDPG \cite{lillicrap2015continuous,plappert2017parameter}, PPO \cite{schulman2017proximal}, and ACKTR \cite{wu2017scalable}. Those algorithms demonstrated learning effective policies in Atari games and physics-based simulations without using any control specific priors. 

Most existing work focuses on training or optimization algorithms, while the policy networks are typically represented as generic multilayer perceptrons (MLPs) for the non-vision part \cite{schulman2015trust, mnih2016asynchronous,salimans2017evolution,schulman2017proximal,wu2017scalable}. There are only few efforts exploring adding specific structures to DRL policy networks, such as inductive bias for improving sampling efficiency, episodic reward, and generalization \cite{wang2016dueling}. 

Some of the recent work also focused on studying network architectures for DRL. The Dueling network of~\cite{wang2016dueling} splits the Q-network into two separate estimators: one for the state value function and the other one for the state-dependent action advantage function. They demonstrated state-of-the-art performance on discrete control of the Atari 2600 domain. However, the dueling Q-network is not easily applicable to continuous control problems. Here, instead of value or Q network, we directly add continuous control-specific structure into the policy network. \cite{tu2017least} studied the sampling efficiency of least-squares temporal difference learning for Linear Quadratic Regulator. Consistent with some of our findings, \cite{rajeswaran2017towards} showed that a linear policy can still achieve reasonable performance on some of the MuJoCo continuous control problems. This also supports the idea of finding a way to effectively incorporate a linear policy into the architecture. 

The idea of splitting nonlinear and linear components can also be found in traditional feedback control liturature \cite{khalil1996noninear}, with successful applications in control of aircrafts \cite{camacho2013model}, robotics, and vehicles \cite{thrun2006stanley}; albeit, those control methods are not learned and are typically mathematically designed through control and stability analysis. Our intuition is inspired by the physical interpretations of those traditional feedback control approaches. 

Similar ideas of using both nonlinear and linear networks have been explored in the vision domain, e.g. ResNet~\cite{he2016deep}, adding linear transformations in perceptrons \cite{raiko2012deep}, and Highway networks \cite{srivastava2015highway}. Our architecture resembles those and probably shares the benefits of easing signal passing and optimization. In addition, we experimentally show the learned sub-modules of our architecture are functional control policies and are robust against both state and action noise.

\section{Background}
\label{background}

We formulate the problem in a standard reinforcement learning setting, where an agent interacts with an infinite-horizon, discounted Markov Decision Process~$(\mathbb{O}, \mathbb{S},\mathbb{A}, \gamma, P, r)$. $\mathbb{S}~\subseteq~\mathbb{R}^n$, $\mathbb{A} \subseteq  \mathbb{R}^m$. At time~$t$, the agent chooses an action $a_t \in \mathbb{A} $ according to its policy $\pi_{\theta}(a_t|s_t)$ given its current observation $o_t \in \mathbb{O}$ or state $s_t \in \mathbb{S}$, where policy network $\pi_{\theta}(a_t|s_t)$ is
parameterized by $\theta$.
For problems with visual input as observation $o_t$, state $s_t$ is considered to be the encoded state after visual processing using a ConvNet, i.e. $s_t = \mu(o_t)$. The environment returns a reward $r(s_t, a_t)$ and transitions to the next state $s_{t+1}$ according to the transition probability $P(s_{t+1}|s_t,a_t)$. The goal of the agent is to maximize the expected $\gamma$-discounted cumulative return: 
\begin{equation} 
J(\theta) = E_{\pi_{\theta}}[R_t] = E_{\pi_{\theta}}[\sum_{i=0}^{\infty}  \gamma^{i} r(s_{t+i}, a_{t+i})] 
\end{equation}
with respect to the policy network parameters $\theta$. 

\section{Structured Control Network Architecture}

\begin{figure*}[t!]
\begin{center}
\includegraphics[trim = 0mm 0mm 0mm 0mm, clip,width=1.7\columnwidth]{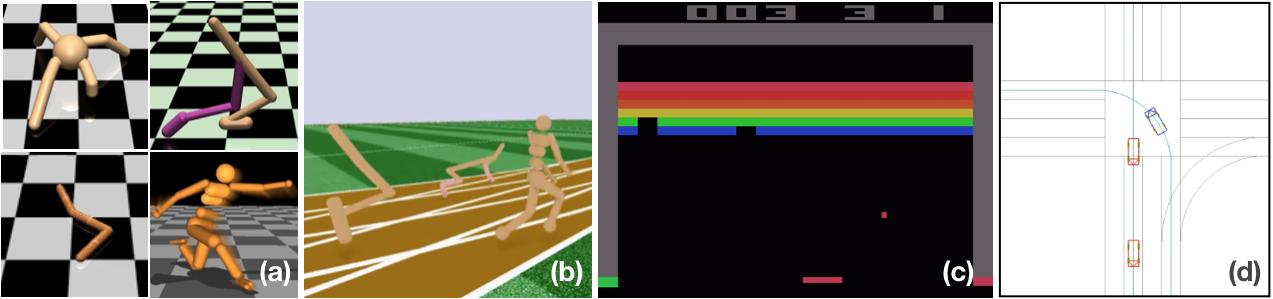}
\caption{\small Various environments: (a) MuJoCo, (b) Roboschool, (c) Atari games, (d) Urban driving environments} \label{fig:envs}
\end{center}
\vspace{-0.25in}
\end{figure*}

In this section, we develop an architecture for policy network  $\pi_{\theta}(a_t|s_t)$ by introducing control-specific priors in its structure. The proposed Structured Control Net (SCN) splits the generic multilayer perceptron (MLP) into two separate streams: a nonlinear control and a linear control. The two streams combine additively into the final action. Specifically, the resulting action at time $t$ is 
\begin{equation} \label{eq:action}
a_t = u^n_t + u^l_t,
\end{equation}
where $u^n_t$ is a nonlinear control module, and $u^l_t$ is a linear control module. Intuitively, the nonlinear control is for forward-looking and global control, while the linear control stabilizes the local dynamics around the residual of global control. The proposed network architecture is shown in Figure~\ref{fig:arch1}. 

This decoupling of control modules is inspired by traditional nonlinear control theory \cite{khalil1996noninear}. To illustrate, let us consider the example of the trajectory tracking problem. In this setting, we typically have the `desired' optimal trajectory of the state, denoted as $s^d_t$, provided by the planning module. The goal of the control is to track the desired trajectory as close as possible, i.e. the tracking error, $e_t = s_t - s^d_t$, should be small. To this end, in the nonlinear control 
setting, the action is given by:
\begin{equation} \label{eq:tracking1}
\begin{aligned}
a_t = u^s_t + u^e_t = u^s_t(s_t, s^d_t) + K \cdot (s_t - s^d_t),
\end{aligned}
\end{equation}
where $u^s_t(s_t,s^d_t)$ is a nonlinear control term, defined as a function of $s_t$ and $s^d_t$, while $u^e_t = K \cdot (s_t - s^d_t)$ is a linear control term, with $K$ being the linear control gain matrix for the tracking error $e_t$. Control theory tells us that the nonlinear term, $u^s_t$, is for global feedback control and also feed-forward compensation based on the predicted system behavior, $s^d_t$, while the linear control, $u^e_t$, is for reactively maintaining the local stability of the residual dynamics of~$e_t$.  At the first glance, Eq.~\ref{eq:tracking1} looks different from Eq.~\ref{eq:action}. However, if we apply the following transformation, $u^n_t(s_t, s^d_t) = u^s_t(s_t, s^d_t) - K \cdot s^d_t $, we obtain:
\begin{equation} \label{eq:tracking2}
\begin{aligned}
a_t & = u^s_t(s_t, s^d_t) + K \cdot (s_t - s^d_t) \\
& = u^s_t(s_t, s^d_t) + K \cdot s_t - K \cdot s^d_t  = u^n_t(s_t, s^d_t) +  K \cdot s_t,
\end{aligned}
\end{equation}
where $u^n_t(s_t, s^d_t) $ can be viewed as the lumped nonlinear control term, and $ u^l_t = K \cdot s_t$ is the corresponding linear control. This formulation shows that the solution for the tracking problem can be converted into the same form as SCN, providing insights into the intuition behind SCN. Moreover, learning the nonlinear module $u^n_t(s_t, s^d_t) $ would imply learning the desired trajectory, $s^d_t$, (planning) implicitly.

\begin{figure}[t]
\begin{center}
\includegraphics[trim = 2mm 2mm 0mm 1mm, clip,width=1.0\columnwidth]{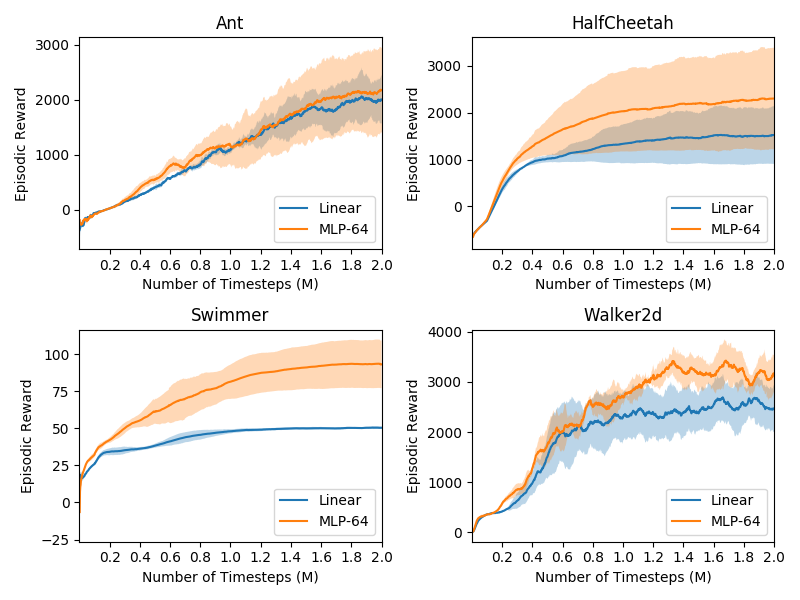}
\caption{\small Example learning curves of MuJoCo environments using Linear policies vs. MLP policies averaged across three trials.} \label{fig:LvsO}
\end{center}
\vspace{-0.1in}
\end{figure}

For the linear control module of SCN, the linear term is $u^l_t = K \cdot s_t +b$, where $K$ is the linear control gain matrix and $b$ is the bias term, both of which are learned. To further motivate the use of linear control for DRL, we empirically observed that a simple linear policy can perform reasonably well on some of the MuJoCo tasks, as shown in Figure~\ref{fig:LvsO}. 

For the nonlinear control module of SCN, we use a standard fully-connected MLP, but remove the last bias term from the output layer, as the bias is provided by the linear control module. In the next section, we show the size of our nonlinear module can be much smaller than standard Deep RL MLP networks due to our split architecture. We also show that both control modules of the architecture are essential to improving model performance and robustness.

\begin{figure*}[t!]
\begin{center}
\includegraphics[trim = 2mm 2mm 1mm 1mm, clip,width=2.0\columnwidth]{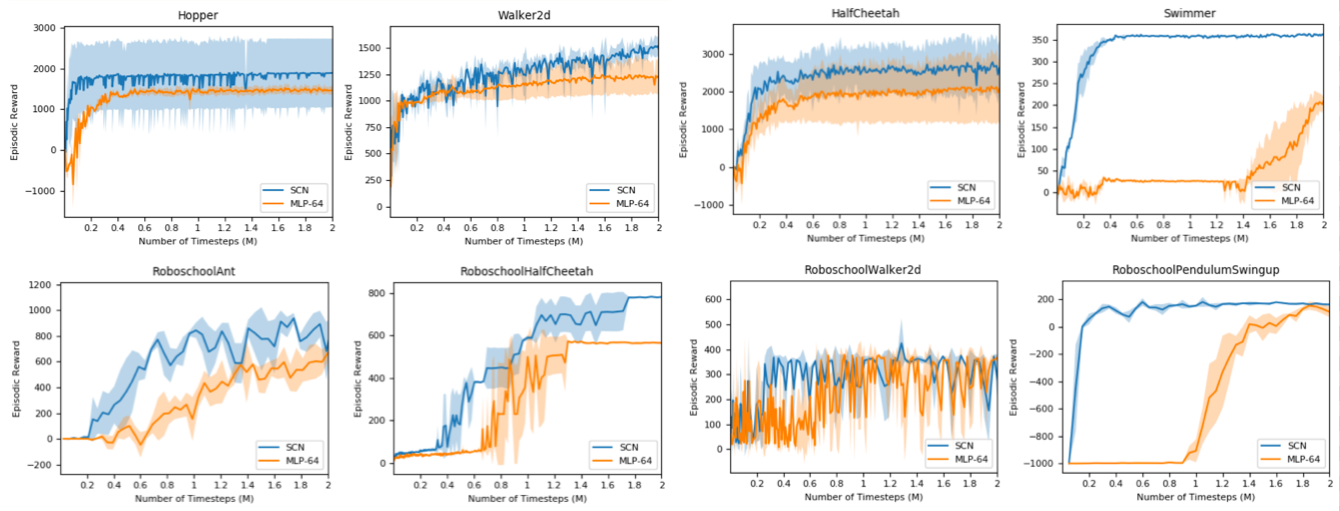}
\caption{\small Learning curves of SCN-16 in blue, and baseline MLP-64 in orange, for ES on MuJoCo and Roboschool environments.} \label{fig:es_results1}
\vspace{-0.1in}
\end{center}
\end{figure*}

\begin{figure*}[t!]
\begin{center}
\includegraphics[trim = 1mm 1mm 1mm 1mm, clip,width=2.0\columnwidth]{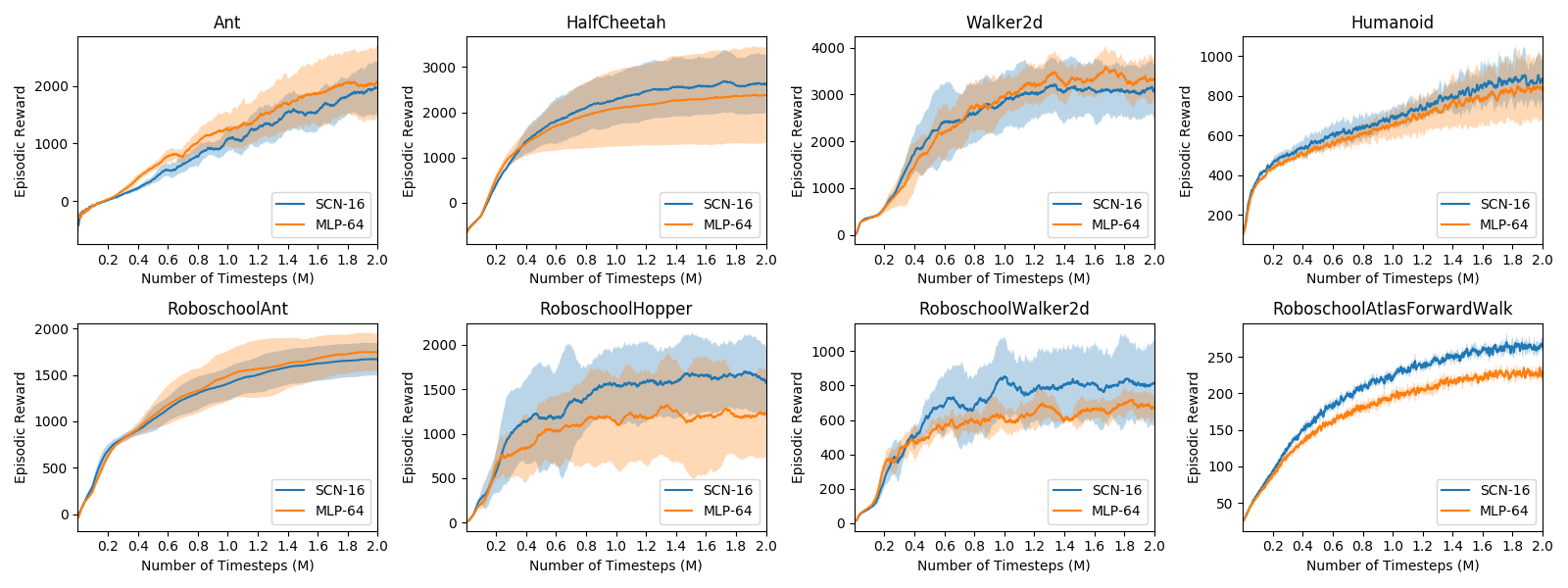}
\caption{Learning curves of SCN-16 (blue), and baseline MLP-64 (orange), for PPO on MuJoCo and Roboschool environments.} \label{fig:ppo_results1}
\end{center}
\vspace{-0.1in}
\end{figure*}

\section{Experimental Setup}

We design and conduct all of our experiments following the guidelines introduced by the recent study on DRL reproduciblity~\cite{henderson2017deep}. To demonstrate the general applicability and effectiveness of our approach, we experiment across different training algorithms and a diverse set of environments. 

\subsection{Environments}

We conduct experiments on several benchmarks, shown in Figure~\ref{fig:envs}, including OpenAI MuJoCo v1 \cite{todorov2012mujoco}, OpenAI Roboschool v1 \cite{roboschool}, and Atari Games~\cite{bellemare2013arcade}. OpenAI Roboschool has several environments similar to those of MuJoCo, but the physics engine and model parameters differ \cite{roboschool}. In addition, we test our method on a custom urban driving simulation that requires precise control and driving negotiations, e.g. yielding and merging in dense traffic. 

\subsection{Training Methods}

We train the proposed SCN using several state-of-the-art training methods, i.e., Evolutionary Strategies (ES) \cite{salimans2017evolution}, PPO \cite{schulman2017proximal}, and ACKTR \cite{wu2017scalable}.

For our ES implementation, we use an efficient shared-memory implementation on a single machine with 48 cores. We set the noise standard deviation and learning rate as 0.1 and 0.01, respectively, and the number of workers to 30. For PPO and ACKTR, we use the same hyper-parameters and algorithm implementation from OpenAI Baselines \cite{baselines}.

We avoid any environment specific hyper-parameter tuning and fixed all hyper-parameters throughout the training session and across experiments. Favoring random seeds has been shown to introduce large bias on the model performance \cite{henderson2017deep}. For fair comparisons, we also avoid any specific random seed selections. As a result, we only varied network architectures based on the experimental need.

\section{Results}

Our primary goal is to empirically investigate if the proposed architecture can bring together the benefits of both linear and nonlinear policies in terms of improving training sampling efficiency, final episodic reward, and generalization of learned policy, while using a smaller network. 

Following the general experimental setup, we conducted seven sets of experiments:

(1) \textbf{Performance of SCN vs. baseline MLP}: we compare our SCN with the baseline MLP architecture (MLP-64) in terms of sampling efficiency, final episodic reward, and network size.

(2) \textbf{Generalization and Robustness}: we compare our SCN with the baseline MLP architecture (MLP-64) in terms of robustness and generalization by injecting action and observation noise at test time.

(3) \textbf{Ablation Study of the SCN Performance}: we show how performance of the Linear policy and 
Nonlinear MLP, used inside of the SCN, compares.

(4) \textbf{Ablation Study of Learned Structures}: we test if SCN has effectively learned functioning linear and nonlinear modules by testing each learned modules in isolation.

(5) \textbf{Performance of Environment-specific SCN vs. MLP}: we compare the best SCN and the best MLP architecture for each environment.

(6) \textbf{Vehicle Driving Domain}: we test the effectiveness of SCN on solving driving negotiation problems from the urban self-driving domain.

(7) \textbf{Atari Domain}: we show the ability of SCN to effectively solve Atari environments.

\subsection{Performance of SCN vs. baseline MLP}

The baseline MLP architecture (MLP-64) used by most previous algorithms \cite{schulman2017proximal,wu2017scalable}, is a fully-connected MLP with two hidden layers, each consisting 
of 64 units, and tanh nonlinearities. For each action dimension, the network outputs the mean of a Gaussian distribution, with variable standard deviation. 
For PPO and ACKTR, the nonlinear module of the SCN is an MLP with two hidden layers, each containing 16 units. For ES, the nonlinear module of SCN is an MLP with a single hidden layer, containing 16 units, with the SCN outputting the actions directly due to the inherent stochasticity of the parameter space exploration.

For each experiment, we trained each network for 2M timesteps and averaged over 5 training runs with random seeds from 1 to 5 to obtain each learning curve. The training results of ES and PPO for MuJoCo/Roboschool are shown in Figure~\ref{fig:es_results1} and Figure~\ref{fig:ppo_results1} respectively. The ACKTR plots are not shown here due to their similarity to PPO. From the results, we can see that the proposed SCN generally performs on par or better compared to baseline MLP-64, in terms of training sampling efficiency and final episodic reward, while using only a fraction of the weights. As an example, for ES, the size of SCN-16 is only 15.6\% of the size of baseline MLP-64 averaged across 6 MuJoCo environments. 

We calculated the training performance improvement to be the percentage improvement for average episodic reward. The average episodic reward is calculated over all~2M timesteps of the corresponding learning curve. This metric indicates the sampling efficiency of training, i.e. how quickly the training progresses in addition to the achieved final reward. Even with the same hidden-layer size, across all the environments, SCN-16 achieved an averaged improvement of 13.2\% for PPO and 17.5\% for ES, compared to baseline MLP-64. 

\begin{figure}[t]
\vspace{-0.1in}
\begin{center}
\includegraphics[trim = 5mm 0mm 0mm 3mm, clip,width=0.7\columnwidth]{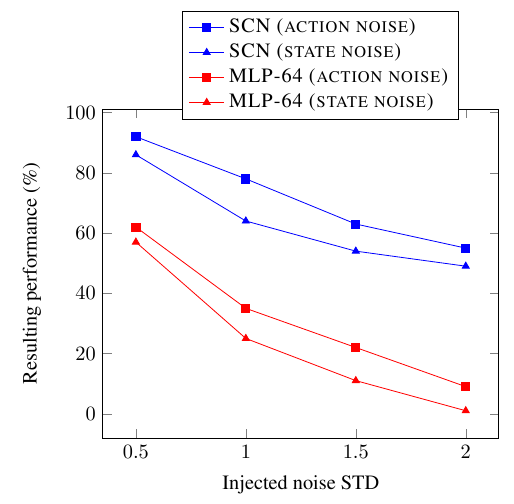}
\caption{\small{Performance degradation of SCN compared to the baseline MLP-64 when varying levels of noise are injected into the action and state space.}} \label{fig:robustness}
\end{center}
\vspace{-0.1in}
\end{figure}

\begin{figure*}[t]
\begin{center}
\includegraphics[trim = 0mm 0mm 0mm 0mm, clip,width=2.0\columnwidth]{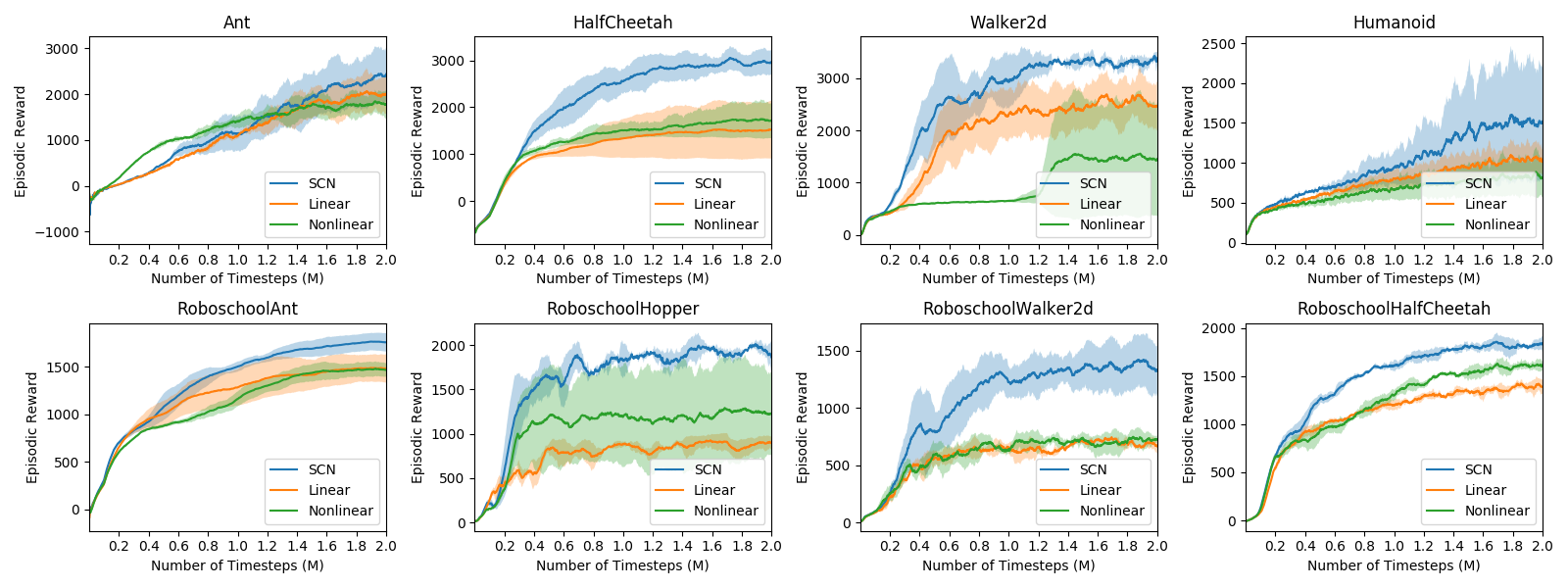}
\caption{\small Ablation study on training performance: SCN in blue, Linear policy in orange, and MLP policy in green (same size as nonlinear module of SCN), trained with PPO on MuJoCo and Roboschool tasks.} \label{fig:ablation}
\end{center}
\vspace{-0.1in}
\end{figure*}

\subsection{Generalization and Robustness}

We next tested generalization and robustness of the SCN policy by injecting action and observation noise at test time, and then comparing with the baseline MLP. We injected varying levels of random noise by adjusting the standard deviation of a normal distribution. The episodic reward for each level of noise is averaged over 10 episodes. Figure~\ref{fig:robustness} shows the superior robustness of the proposed SCN against noise unseen during training compared to the baseline MLP-64. 

\subsection{Ablation Study of the SCN Performance}

To demonstrate the synergy between the linear and the nonlinear control modules for SCN architecture, we trained the different sub-modules of SCN separately and conducted ablation comparison, i.e. linear policy and nonlinear MLP policy with the same size as the nonlinear module of SCN. The results on the MuJoCo and OpenAI Roboschool with PPO are shown in Figure~\ref{fig:ablation}. We make following observations: (1) the linear policy alone can be trained to solve the task to a reasonable degree despite its simplicity; (2) SCN outperforms both the linear and nonlinear MLP policy alone by a large margin for most of the environments.

\subsection{Ablation Study of Learned Structures}

To investigate whether SCN has learned effective linear and nonlinear modules, we conducted three tests after training the model for 2M timesteps. Here, we denote the linear module inside SCN as SCN-L and nonlinear module inside SCN as SCN-N to distinguish from the separately trained Linear policy and ablation MLP policy. We run SCN-L or SCN-N by simply not using the other stream.

The first test compares the performance of the separately trained Linear policy, with the linear control module learned inside SCN (SCN-L). In simpler environments, where a Linear policy performs well, SCN-L performs similarly. Thus SCN appears to have learned an effective linear policy. However, in more complex environments, like Humanoid, SCN-L does not perform well on its own emphasizing the fact that the nonlinear module is very important. Across MuJoCo environments, SCN-L is able to achieve 68\% of the performance of the stand alone Linear policy when trained with ES, and 65\% with PPO. Hence for most environments, the linear control of SCN is effectively learned and functional.

\begin{table}[t]
\small
\begin{center}
\begin{tabular}{p{2.0cm}| p{1.0cm} p{1.0cm}| p{1.0cm} p{1.0cm}}
\hline
        & Final & Final & Average & Average \\ \hline
Task    & SCN & MLP & SCN & MLP \\ \hline
HalfCheetah & 3310 & \textbf{3386}  & \textbf{2148} & 1693 \\ \hline
Hopper & 2479 & \textbf{2619} & 1936 & \textbf{1977} \\ \hline
Humanoid & \textbf{2222} & 1078 & \textbf{979} & 674 \\ \hline
Walker2d & \textbf{3761} & 3520 & \textbf{2404} & 2349 \\ \hline
Swimmer & 81 & \textbf{109} & 58 & \textbf{74} \\ \hline
Ant & \textbf{2977} & 2948 & \textbf{1195} & 1155 \\ \hline
Roboschool HalfCheetah & 1901 & \textbf{2259} & 1419 & \textbf{1693} \\ \hline
Roboschool Hopper & \textbf{2027} & 1411 & \textbf{1608} & 914 \\ \hline
Roboschool Humanoid & \textbf{187} & 175 & \textbf{131} & 115 \\ \hline
Roboschool Walker2d & \textbf{1547} & 774 & \textbf{1048} & 584 \\ \hline
Roboschool Ant & 1968 & \textbf{2018} & \textbf{1481} & 1394 \\ \hline
Roboschool AtlasForwardWalk & \textbf{273} & 236 & \textbf{202} & 176 \\ \hline
\end{tabular}
\end{center}
\caption{\label{tab:tuned} \small{Results of final episodic reward and averaged episodic reward for best SCN vs. best MLP per environment.}}
\vspace{-0.1in}
\end{table}

The second test compares the performance of the MLP versus the nonlinear module inside SCN (SCN-N). Unlike the linear module test, the performance of the two identical MLPs are drastically different. Across all environments, SCN-N is not able to perform well without the addition of the linear module. We found that SCN-N is only able to achieve about 9\% of the performance of the
stand-alone  MLP when trained with ES, and 8\% with PPO. These tests verify the hypothesis that the linear and nonlinear modules learn very different behaviors when trained in unison as SCN and rely on the synergy between each other to have good overall performance.

The third test compares the performance of SCN versus a pseudo SCN, which is assembled post-training by a pre-trained MLP and a pre-trained Linear policy. For tested environments, the naive combination of the two already-trained policies does not perform well. By combining the separate MLP and linear model, we were able to achieve only 18\% of the performance of SCN when using ES, and 21\% with PPO. This demonstrates the importance of training both components of the structure in the same network, like SCN.

\subsection{Performance of Environment-specific SCN}

In general, different environments have different complexities. To study the SCN for the most efficient size for each environment, we sweep the hidden-layer size of the nonlinear module across the set of model sizes 64, 32, 16, 8, and 4. As a comparison, we also sweep the hidden-layer size of the baseline MLP to get the best MLP size for each environment for the same size set. We keep the number of hidden layers fixed at two.

For each environment, we compared the environment-wise best SCN and best MLP from the model size set. We calculated the average episodic reward as episodic rewards averaged over the whole 2M timesteps of the corresponding learning curve. This metric indicates the sampling efficiency of training, i.e. how quickly the training progresses. Final episodic reward is the averaged rewards of the last 100 episodes. We illustrate the results with data trained with PPO. From Table~\ref{tab:tuned}, we can see SCN shows equal or better performance, compared to the environment-wise best performing MLP.

\begin{figure}[t]
\begin{center}
\includegraphics[trim = 0mm 0mm 0mm 0mm, clip,width=1.0\columnwidth]{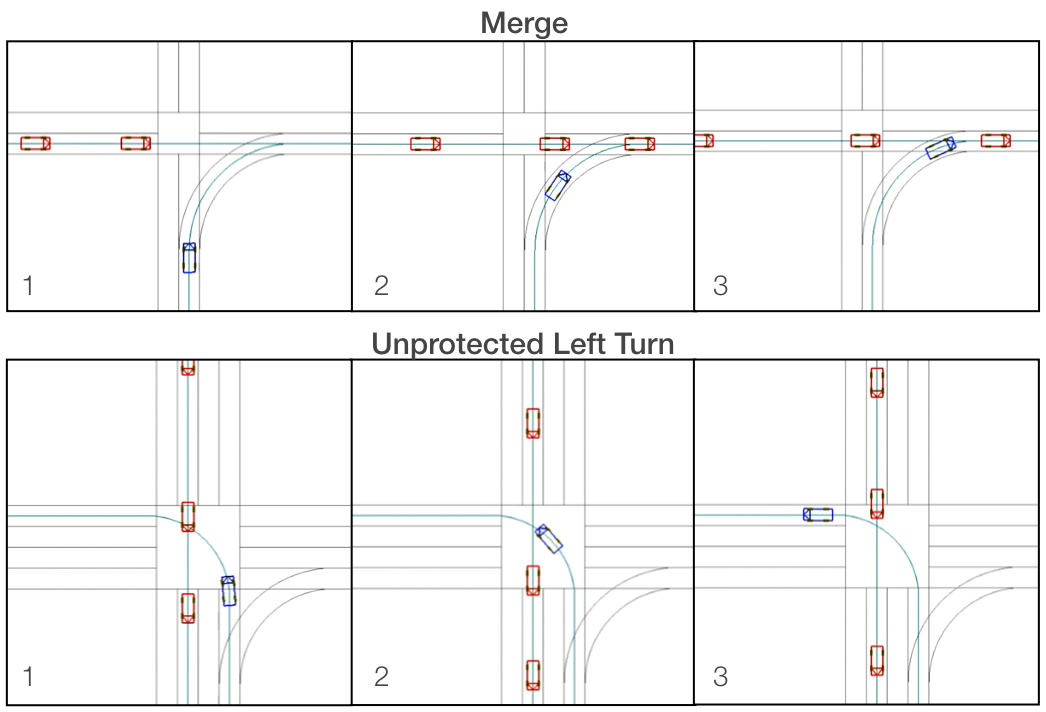}
\caption{\small Two sequences of 3 frames showing learned agent (blue) via SCN, making a merge (top 3) and an unprotected left turn (bottom 3) with oncoming traffic (red).} \label{fig:unprotectedleft_merge}
\end{center}
\vspace{-0.1in}
\end{figure}

\subsection{Vehicle Driving Domain}

\begin{table}[t]
\begin{center}
\textbf{Urban Driving Results} \\	
\small
\begin{tabular}{p{3cm} p{1.5 cm} p{1.5 cm} }
\hline
Task	& ES SCN/MLP  & PPO SCN/MLP \\ \hline
UnprotectedLeftTurn & 93/76 & 138/102 \\ \hline
Merge & 95/81 & 101/88 \\ \hline
\end{tabular}
\end{center}
\caption{\label{tab:results6} \small{Final episodic rewards on the driving scenarios using the SCN vs. MLP with ES and PPO.}}
\vspace{-0.1in}
\end{table}

We next validate the effectiveness of SCN on solving negotiation problems in the urban self-driving domain. Sequential decisions in dense traffic are difficult for human drivers. We picked two difficult driving scenarios: completing an unprotected left turn and learning to merge in dense traffic.

For the simulation, we used a bicycle model as the vehicle dynamics~\cite{thrun2006stanley}. Simulation updates at 10Hz. The other traffic agents are driven by an intelligent driver model with capabilities of adaptive cruise control and lane keeping. Both learned agents and other traffic agents are initialized randomly within a region and a range of starting speeds. The other traffic agents have noise injected into their distance keeping and actions for realism. The state observed by the agent consists of ego vehicle state, states of other traffic agents, and the track on which it is traveling (e.g. center lane). The reward is defined to be -200 for a crash, 200 for reaching the goal, small penalties for being too close to other agents, small penalties for going too slow or too fast, and small incentives for making progress on the track. An episode reward larger than 50 is considered to be solved. A Stanley steering controller \cite{thrun2006stanley} is used to steer the vehicle to follow the track. The action space for the learned agent is a continuous acceleration value. 

Table~\ref{tab:results6} shows the final episodic reward achieved, comparing SCN and MLP-64, trained with ES and PPO. In Figure~\ref{fig:unprotectedleft_merge}, we visualize the learned SCN policies controlling the learned agent (blue) while successfully making a merge and an unprotected left turn through a traffic opening. 

\subsection{Atari Domain}

\begin{figure*}[t!]
\centering
\includegraphics[trim = 0mm 0mm 0mm 0mm, clip,width=2.0\columnwidth]{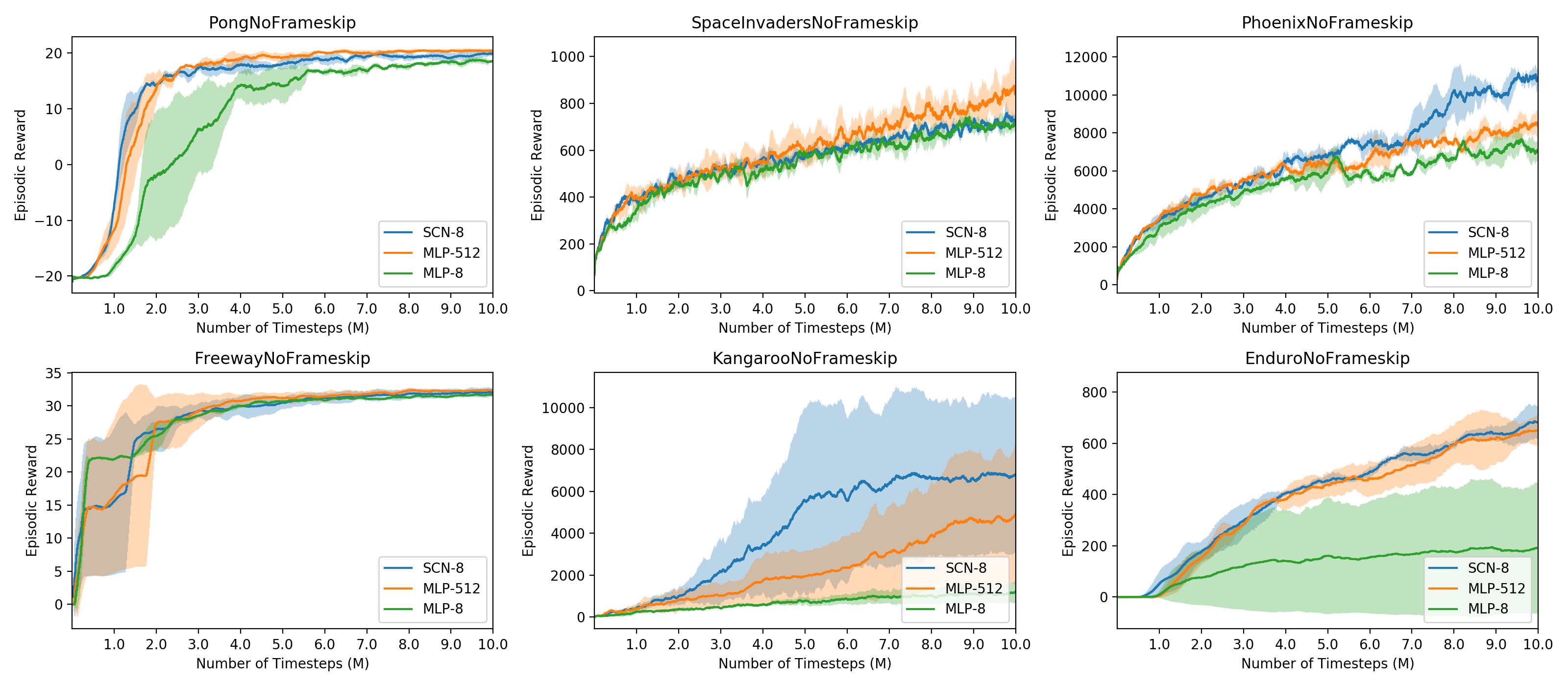}
\caption{\label{fig:atari_results} \small Atari environments: SCN-8 in blue, baseline Atari MLP (MLP-512) in orange, and ablation MLP (MLP-8) in green. 10M timesteps equal 40M frames.}
\vspace{-0.1in}
\end{figure*}

\begin{table}[t]
\begin{center}
\small
\textbf{Atari Results} \\	
\begin{tabular}{c c c }
\hline
Metric	& SCN  & MLP \\ \hline
Average Reward & \textbf{34} & 26 \\ \hline
Final Reward & \textbf{35} &  25 \\ \hline
\end{tabular}
\end{center}
\vspace{-0.1in}
\caption{\label{tab:atari} \small{Number of Atari games won by SCN-8 vs. MLP-512 when trained with PPO.}} 
\vspace{-0.1in}
\end{table}

In this section, we show that SCN is able to learn effective policies in Atari environments. The SCN with visual inputs uses the same convolutional layers and critic model as \cite{schulman2017proximal}, but the learned visual features are flattened and fed into the linear and nonlinear modules of SCN. The baseline Atari policy from PPO \cite{schulman2017proximal} is a fully-connected MLP with one hidden layer containing 512 units, chosen by cross-validation, on the flattened visual features, which we denote as MLP-512. The nonlinear module of SCN is a MLP with one hidden layer that has 8 units (SCN-8).

Each learning curve and metric is averaged across three 10M-timestep trials (random seeds: 0,1,2).  Learning curves for all 60 games are provided in the Appendix. We summarize the results in Table~\ref{tab:atari}, where we show that the SCN-8 can perform competitively in comparison to Atari baseline policy, MLP-512 (much larger in size), across 60 Atari games. If the metric is similar, we consider SCN wins since it is smaller in size.
Figure~\ref{fig:atari_results} displays learning curves for 6 randomly chosen games. 
We can see that SCN-8, achieves equal or better learning performance compared to the baseline policy, MLP-512, and the ablation policy, MLP-8.
We further observed that even with 4 hidden units, SCN-4, which is smaller in size than SCN-8, performs similarly well on many games tested.

\section{Case Study: Locomotion-specific SCN}

In our final set of experiments, we use dynamic legged locomotion as a case study to demonstrate how to tailor SCN to specific tasks using the task-specific priors.

\begin{figure}[t]
\centering
\includegraphics[trim = 0mm 0mm 0mm 0mm, clip,width=1.0\columnwidth]{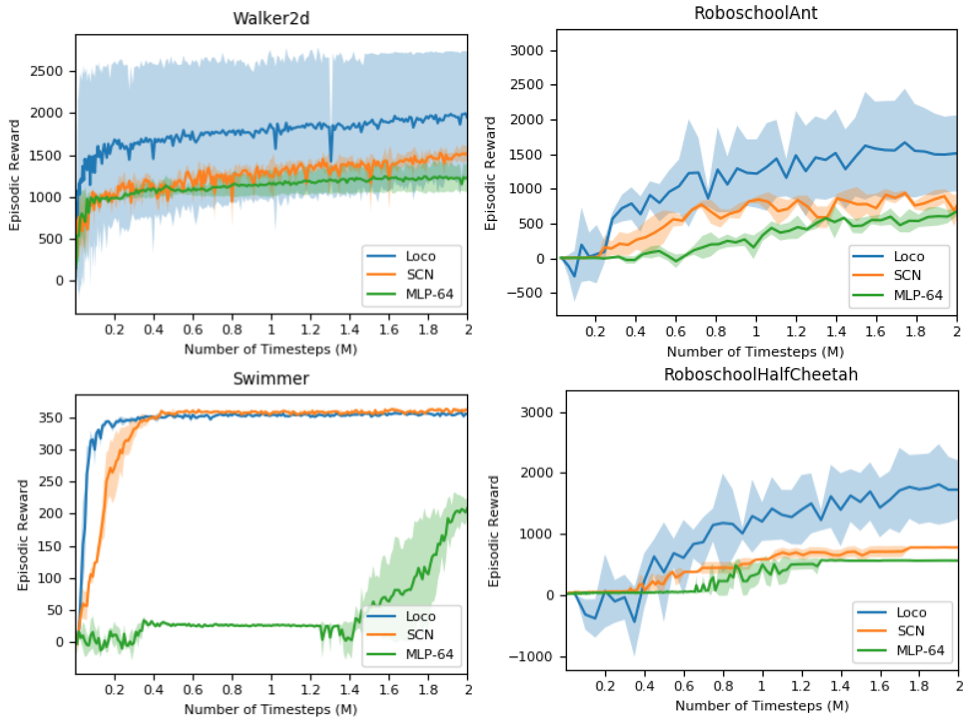}
\caption{\label{fig:locomotor_results} Locomotive tasks from MuJoCo and Roboschool: locomotor net (Loco) in blue, SCN in orange, and baseline MLP (MLP-64) in green. Results achieved using ES.}
\end{figure}

In nature, neural controllers for locomotion have specific structures, termed central pattern generators (CPGs), which are neural circuits capable of producing coordinated rhythmic patterns \cite{ijspeert2008central}. While the rhythmic motions are typically difficult to learn with general feedforward networks, by emulating biological CPGs using Fourier series and training the Fourier coefficients, we are able to show the power of adding this inductive bias when learning cyclic movements in the locomotive control domain. The nonlinear module of SCN becomes 
\vspace{-0.1in}
\begin{equation}
u^n_t =\sum\limits_{i=1}^cA_isin(\omega_it + \phi_i), 
\end{equation}
where for each action dimension, $A_i$, $\omega_i$, $\phi_i$ are the amplitude, frequency and phase, respectively, of the component~$i$, that will be learned,
and $c$ is set to $16$ sinusoids. In our experimental results, we find that the linear module of SCN is necessary for achieving better performance, by stabilizing the system around the residual of CPGs outputs. We name this specific instantiation of SCN ``Locomotor Net''. By replacing the MLP in the nonlinear module of SCN with a locomotive-specific implementation, we were able to further improve sampling efficiency and the final reward on those environments. Example results on locomotive tasks from MuJoCo and Roboschool are shown in Figure~\ref{fig:locomotor_results}.

\section{Conclusion}

In this paper we developed a novel policy network architecture that is simple, yet effective. The proposed Structured Control Net (SCN) splits the generic MLP into two separate streams: a nonlinear control module and a linear control module. We tested SCN across 3 types of training methods (ES, PPO, ACKTR) and 4 types of environments (MuJoCo, Roboschool, Atari, and simulated urban driving), with various ablation and generalization tests. We experimentally demonstrated the benefits of both linear and nonlinear policies: improving training sampling efficiency, final episodic reward, and generalization of learned policy, in addition to using a smaller network and being general and applicable to different training methods. By incorporating problem specific priors into the architecture, the proposed architecture has the potential to improve upon broader control tasks. Our case study demonstrated much improved performance for locomotion tasks, by emulating the biological central pattern generators (CPGs) as the nonlinear part of the architecture. For future work, we plan to extend the SCN to incorporate a planning module. The planning module will be responsible for long-term planning and high-level abstracted decision making.

\subsubsection*{Acknowledgments}

We thank Emilio Parisotto, Yichuan Tang, Nitish Srivastava, and Hanlin Goh for helpful comments and discussions. We also thank Russ Webb, Jerremy Holland, Barry Theobald, and Megan Maher for helpful feedback on the manuscript.

\bibliographystyle{ieee}

\newpage

\section*{}
\newpage
\appendix
\section*{Appendix A: Atari Results}
\noindent
\begin{minipage}{1.0\textwidth}
\centering
	\includegraphics[height=0.9\vsize]{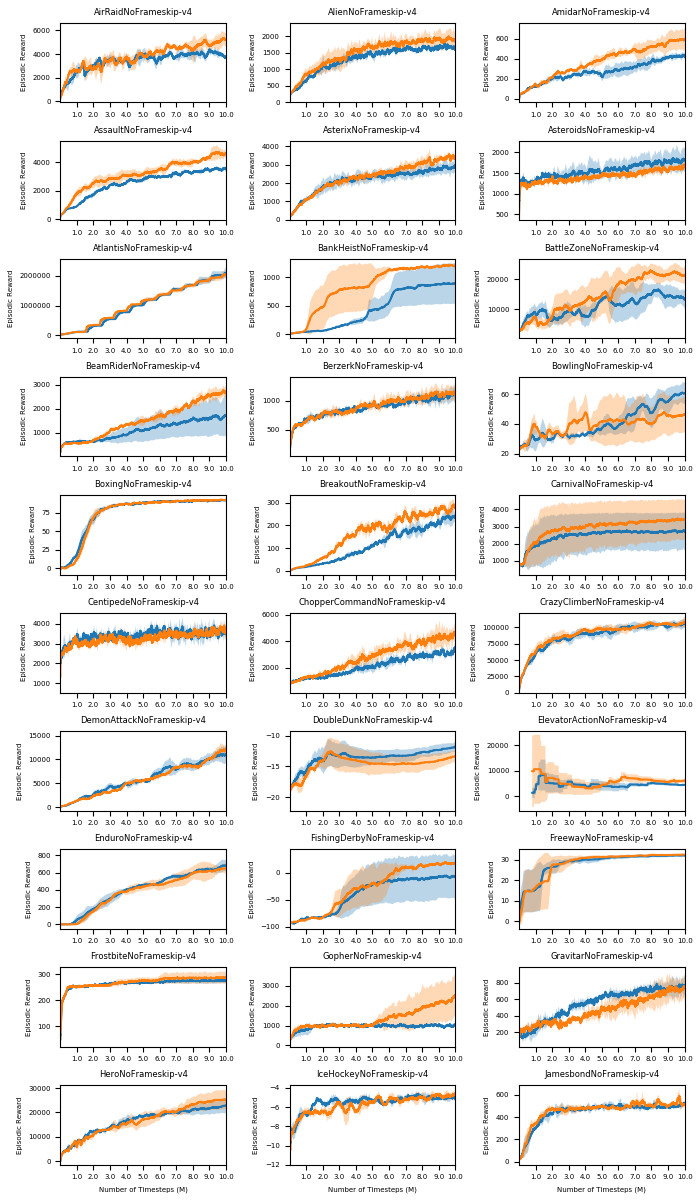}
\captionof{figure}{Comparison of SCN-8 (blue) and MLP-512 (orange) on number 1-30 of all 60 Atari games included in OpenAI Gym at the time of publication.} \label{fig:atari_all1} 
\end{minipage}

\begin{figure*}[!ht]
\centering
\includegraphics[trim = 1mm 1mm 1mm 1mm, clip,width=1.5\columnwidth]{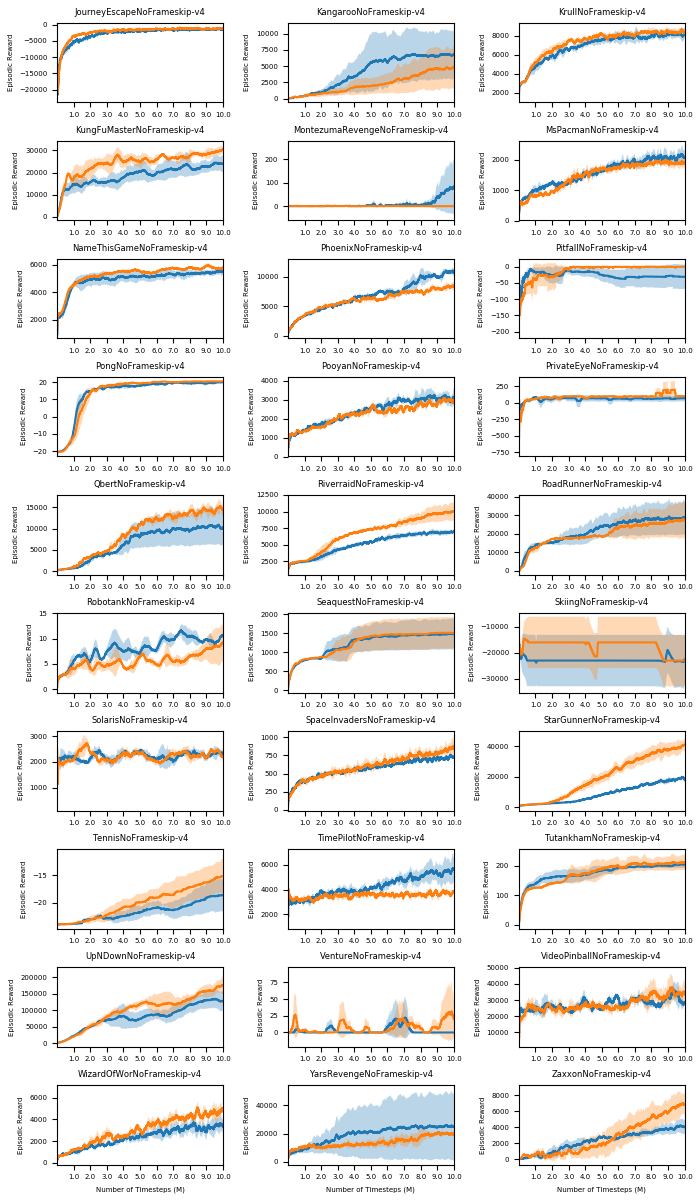}
\caption{\label{fig:atari_all2} Comparison of SCN-8 (blue) and MLP-512 (orange) on number 31-60 of all 60 Atari games included in OpenAI Gym at the time of publication.}
\end{figure*}

\end{document}